\documentclass[runningheads]{llncs}

\usepackage{graphicx}
\usepackage{amsmath,amssymb}
\usepackage{multirow}
\usepackage{color}
\usepackage{booktabs}
\usepackage{subfig}
\usepackage{url}

\begin{document}
\title{Simultaneous Recognition of Horizontal and Vertical Text in Natural Images}
\titlerunning{Recognition of Horizontal and Vertical Text}

\author{Chankyu Choi, Youngmin Yoon, Junsu Lee, Junseok Kim}

\authorrunning{C. Choi et al.}

\institute{NAVER Corporation\\
\email{\{chankyu.choi,youngmin.yoon,junsu.lee,jun.seok\}@navercorp.com}}

\maketitle
\begin{abstract}
Recent state-of-the-art scene text recognition methods have primarily focused on horizontal text in images. However, in several Asian countries, including China, large amounts of text in signs, books, and TV commercials are vertically directed. Because the horizontal and vertical texts exhibit different characteristics, developing an algorithm that can simultaneously recognize both types of text in real environments is necessary. To address this problem, we adopted the direction encoding mask (DEM) and selective attention network (SAN) methods based on supervised learning. DEM contains directional information to compensate in cases that lack text direction; therefore, our network is trained using this information to handle the vertical text. The SAN method is designed to work individually for both types of text. To train the network to recognize both types of text and to evaluate the effectiveness of the designed model, we prepared a new synthetic vertical text dataset  and collected an actual vertical text dataset (VTD142) from the Web. Using these datasets, we proved that our proposed model can accurately recognize both vertical and horizontal text and can achieve state-of-the-art results in experiments using benchmark datasets, including the street view test (SVT), IIIT-5k, and ICDAR. Although our model is relatively simple as compared to its predecessors, it maintains the accuracy and is trained in an end-to-end manner.
\end{abstract}

\section{Introduction}
Optical character recognition (OCR) in the wild is a challenging problem in computer-vision research, and several approaches have been studied to perform OCR on images acquired from various environments such as documents or natural scenes. OCR is usually segmented into separate modules such as scene text detection and scene text recognition. The scene text detection module identifies the regions of interest in images that contain text; various algorithms handle this task. Subsequently, the scene text recognition module translates these regions into labeled text. This study presents a novel method to perform scene text recognition.

\begin{figure}[tbp]
\centering
\centerline{\includegraphics[width=0.9\linewidth]{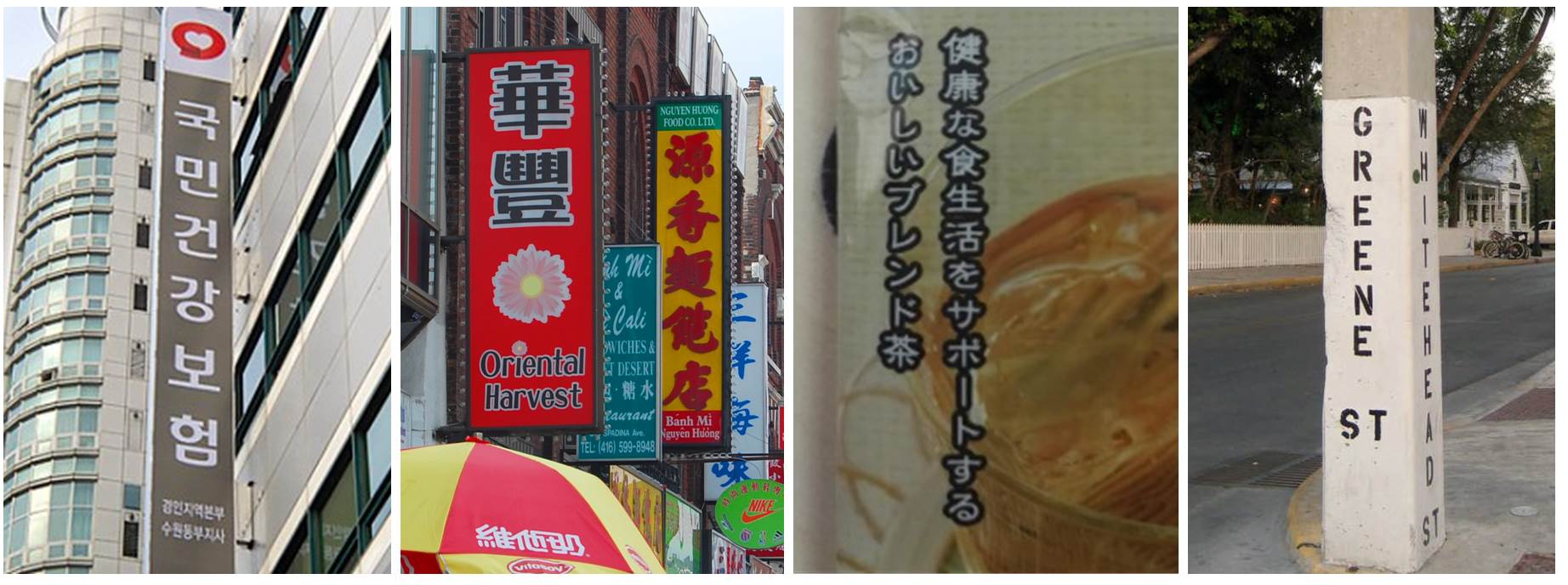}}
\caption{Majority of the prior scene text recognition algorithms only attempted to cover the horizontal text. However, in several Asian countries, including China, large amounts of text in signs, books, and TV commercials are vertically directed.
}
\label{vertical_text}
\end{figure}

Majority of the prior scene text recognition algorithms only attempted to cover the horizontal text. However, in several Asian countries, including China, large amounts of text in signs, books, and TV commercials are vertically oriented. Horizontal and vertical texts exhibit different characteristics that should be simultaneously solved for. They occupied different aspect ratio text region in image. Therefore, a clever algorithm must be developed to deal with both types of text. Fig.\ref{vertical_text} depicts the reason due to which the OCR algorithm is able to handle the vertically directed text in real environments. A recent study used arbitrarily oriented text recognition (AON)\cite{Zhanzhan2017Arbitrarily} encoded image input in four directions (top-to-bottom, bottom-to-top, left-to-right, and right-to-left) to detect a freely rotated word. AON can, therefore, adapt in an unsupervised manner to tackle the rotated text using four orientations of directional information. 

In this study, inspired by the AON method, we simultaneously recognize the horizontal and vertical texts. Our method uses carefully designed information formulated via the directional encoding mask (DEM) and selective attention network (SAN) methods. Both the modules help the network to learn quickly and accurately. DEM is designed to extract the direction of a text region before it is passed to the text recognition model. Therefore, it should be created in advance during the preparation time of the ground truth data. It encapsulates the directional information for the text region. An advantage of DEM is that it helps the network to quickly and accurately learn to distinguish between vertical and horizontal text in a supervised manner. Our contributions are as follows. 
\begin{itemize}
\item[]1) We created two alternative methods, namely, DEM and SAN that can simultaneously deal with both horizontal and vertical characters.
\item[]2) To train for vertical text, we generated a synthetic vertical text dataset (SVTD). In addition, to evaluate the performance with respect to the vertical text, we collected an actual vertical text dataset (VTD142), including building signs and titles. All the real data were collected from the webpages.  
\item[]3) Our model achieves state-of-the-art results with respect to various benchmark datasets.
\end{itemize}

\section{Related Work}
Previously, the development of OCR algorithms mostly used handcrafted features, which are difficult to design and exhibit limited performance in specific areas; however, currently, neural networks offer a promising alternative for powerful OCR algorithms\cite{krizhevsky2012imagenet,szegedy2015going,he2016deep,he2016identity}. Convolutional recurrent neural network (CRNN)\cite{shi2017end} encode the input images using convolutional neural network (CNN) and further decode the encoded features using recurrent neural network (RNN) by connectionist temporal classification (CTC). Recurrent and recursive convolution network\cite{liang2015recurrent,lee2016recursive} improve the representational power and efficiency of the encoders. Attention OCR\cite{wojna2017attention} improves the overall performance using positional encoding and attention mechanisms. Gated recurrent convolutional neural network (GRCN)\cite{wang2017gated} adopt gated recurrent CNNs to further improve the performance. Currently, even though there are several deep neural network (DNN)\cite{Zhanzhan2017Arbitrarily,Zhanzhan2017Focusing,Baoguang2016Robust} in use, only a few studies that focus on the vertical and horizontal texts have been conducted. FOTS\cite{liu2018fots} handles oriented text, which is essentially different from recognizing characters the in vertical direction. The results also exhibited lower performance as compared to the attention-based decoder because they have experimented with a CTC-based decoder. AON\cite{Zhanzhan2017Arbitrarily} is an unsupervised learning method that learns a character’s direction by network oneself. Conversely, we examine the orientation of a word during the training stage and add this feature to the network so that the network can be learned in a supervised manner. We extended the AON to a supervised learning method to recognize the horizontal and vertical texts.

\section{Proposed Method}
We begin by explaining the manner in which a unified network can be designed to simultaneously handle the horizontal and vertical texts using an image’s aspect ratio. We examine the direction of a word during the training stage and add this feature to the network so that the network can be learned in a supervised manner. To explain it more clearly, our methods determine the direction of the text after checking the width-to-height ratio of the input image. Further, the vertical text is rotated by 90° and input to the network. In this study, we introduce the two methods that are required to achieve this.

\begin{figure}[tbp]
\centering
\centerline{\includegraphics[width=0.9\linewidth]{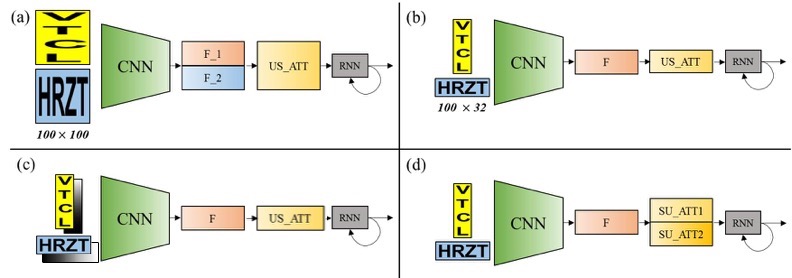}}
\caption{Various models for recognizing the horizontal and vertical text : (a) AON style\cite{Zhanzhan2017Arbitrarily} with 2-directional clue network), (b) Baseline, (c) DEM, (d) SAN. \textit{F} means features, \textit{SU\_ATT} means supervised attention and \textit{US\_ATT} means unsupervised attention mechanism respectively.}
\label{various_models}
\end{figure}

\subsection{Directional Encoding Mask}
DEM combines the input image and the directional information of the characters, which are further encoded by the CNN encoder. Because DEM contains the directional information of the text, the network uses this information while training. In this study, we encode the directional information in cosine for the horizontal text and in sine for the vertical text. The equation below describes the DEM. 
\begin{equation} \label{eq:1}
\text{DEM} = \begin{cases}
	\{y \mid y = \sin(0.5 * normalized\_width * \pi)\}, & \text{if width $>$ height}\\
    \{y \mid y = \cos(0.5 * normalized\_width * \pi)\}, & \text{otherwise}
	\end{cases}
\end{equation}
In the equation(\ref{eq:1}), the normalized width is the x coordinate in normalized space [0,1]. As can be observed from the equation, we encode DEM for the horizontal text using a cosine function. The DEM for vertical text is encoded using a sine function. The kernel function of DEM can be replaced by similar functions; therefore, we chose it experimentally. 
\begin{equation} \label{eq:2}
X = concat(I, DEM)
\end{equation}
DEM is not a single value, but a two-dimensional image of the same size as the input image. DEM and the input image are concatenated and entered into the CNN. $X$ is used as the input of the CNN.

\subsection{Selective Attention Network}
SAN is another supervised learning method. It has two attention masks, each of which reacts according to the corresponding text direction; therefore, they work separately. SAN can turn its own attention weight on and off depending on the direction of the text. This differs from AON’s clue network because it is based on supervised learning. The \textit{character feature vector} for the RNN is defined as :
\begin{equation} \label{eq:3} 
{x^c}_t = W_cc_{t-1}^{OneHot} 
\end{equation}
where $c_{t-1}$ is one hot encoding of the previous letter (ground truth during training and predicted during test time). 

Let $s_t$ be the hidden state of the LSTM at time $t$. Further, we compute the output and the subsequent state of the LSTM, as shown
\begin{equation} \label{eq:4}
(y_t,s_t) = LSTM_{step}(x_t,s_{t-1})
\end{equation}

Let us denote the cnn features as $f$ = {$f_{i,j,c}$} where $i$, $j$ are the index locations in the feature map and $c$ indexes the channels. We verified the aspect ratios of the image inputs and created two attention modules. These attention masks are denoted as ${\alpha}_t^{(h, v)}$, where $h$ and $v$ indicate the horizontal and vertical texts, respectively. We name this as \textit{selective attention network} because it responds to the text direction. 

\begin{equation} \label{eq:5} 
{\alpha}_t^{(h, v)}=softmax(V_a^T\:{\circ}\:tanh(W_s^{(h,v)}s_t+W_ff_{i,j}))
\end{equation}
SAN comprises only one set of attention weights for each horizontal and vertical text. In the equation(\ref{eq:5}), ${\circ}$ is the Hadamard product and ${tanh}$ is applied element-wise to its vector argument. Further, the \textit{visual feature vector} (also known as the \textit{context vector}), ${x^v}_t$ is computed as the weighted image features based on the attention at time step $t$ as shown : 
\begin{equation} \label{eq:6} 
{x^v}_t=\sum_{i,j}{{\alpha}_{t,i,j}}^{(h,v)}f_{i,j,c}
\end{equation}
We can define $x_t$ as the input of LSTM, as depicted bellow.
\begin{equation} \label{eq:6} 
x_t = {x^c}_t + {x^v}_t
\end{equation}
The final predicted distribution over the letters at time $t$ is described as shown. 
\begin{equation} \label{eq:7} 
\tilde{y_t}=W_{\tilde{o}}y_t+b_{\tilde{o}}
\end{equation}
\begin{equation} \label{eq:8} 
\tilde{Y}=\left\{ \tilde{y_0},...,\tilde{y_t},...,\tilde{y_{T-1}} \right\}
\end{equation} 
In the equations, $T$ is the sequence length of the RNN and $\tilde{Y}$ is the result of our LSTM decoder.

\section{Experiments}
\begin{figure}[tbp]
\centering
\centerline{\includegraphics[width=0.9\linewidth]{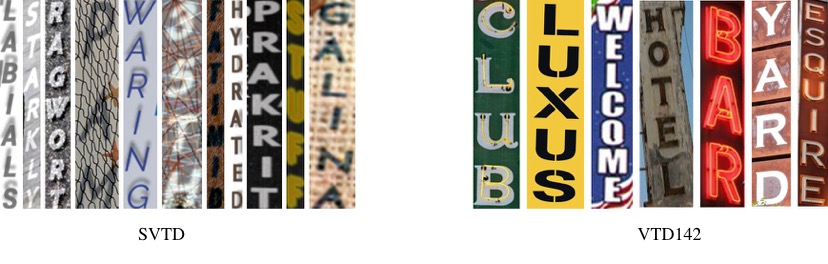}}
\caption{SVTD is a new synthetic vertical text dataset in a similar fashion as [1]. VTD142 is real vertical text dataset, which was collected from web.
}
\label{vertical_dataset}
\end{figure}

\subsection{Dataset}
A range of datasets of images of words are used to train and test the model.

SVT\cite{wang2011end} includes cropped images, including 257 training images and 647 test images, captured from Google Street View

IIIT5k\cite{mishra2012scene} contains 2000 cropped training images and 3000 testing images downloaded from the Google image search engine

ICDAR 2003\cite{lucas2003ICDAR} (IC03 in short) contains 258 full-scene and 936 cropped images for training, and 251 full-scene and 860 cropped images for testing.

ICDAR 2013\cite{Karatzas2013ICDAR} (IC13 in short) contains 1015 cropped text images, but no lexicon is associated.

ICDAR 2015\cite{karatzas2015icdar} (IC15 in short) contains 2077 cropped text images without lexicon.

SVTD is a new synthetic vertical text dataset in a similar fashion as [1]. It includes 1 million training word images and 3,000 testing word images.

VTD142 is real vertical text dataset including 142 word images. We collect it from web.

Large amounts of datasets are needed to create a good-performance, deep-learning model. However, producing annotated training data requires considerable time and effort. Compared to other tasks in computer vision, generating high-quality synthetic data using various font files in optical character recognition (OCR) is easy. In previous studies, millions of synthetic images were created for recognition\cite{JaderbergSVZ14} and they have been further developed for use in detectors\cite{gupta2016synthetic}. However, those studies only dealt with horizontal characters, not vertical ones, which could not be directly used. We applied the method described in \cite{gupta2016synthetic} to find the area in which the letters were rendered. The depth and segmentation maps were estimated from the background image to be synthesized, and the vertical characters were rendered in the homogeneous regions. We also applied augmentation techniques, such as affine transformations, contrast changes, and certain noises, to generate various images. To our knowledge, there are no publicly available vertical-text datasets that can be used for evaluation; hence, we had to collect new data to properly evaluate our proposed method. To this end, we chose several appropriate keywords (e.g., vertical sign, building’s sign, and vertical text) and used them in the web. We usually gathered vertical signs on the buildings because that is the way they are mostly seen in real environments, such as streets, shops, and buildings.

We follow the evaluation protocol of \cite{wang2011end}, which is set to recognize words, that include only alphanumeric characters (0-9 and A-Z) and are at least three characters long. Results are not case-sensitive.

\subsection{Exploring various models}
To recognize the vertical and horizontal texts, we consider various models in Fig.\ref{various_models}. (a) depicts a network based on AON. It is an unsupervised method that learns the text direction without using any relevant information. The input image has a square ratio and is encoded with respect to two directions (top-to-bottom and left-to-right). The clue network (CN) identifies the text direction and focuses on that direction. (b) depicts an unsupervised method, which is our baseline network. It differs from AON because the input image is observed to be rectangular. We hypothesized that a rectangle would be more efficient than a square because we only need to consider two directions, i.e., horizontal and vertical. In fact, the vertical text images are transposed. This hypothesis was proved via experiments that depicted an approximate performance improvement of 2--3$\%$p. (c) depicts that DEM (S), our proposed supervised module, provides the text direction to the network. (d) depicts that SAN, our second proposed supervised network, works selectively depending on the text direction.

\begin{figure}[!t]
\centering
\centerline{\includegraphics[width=0.9\linewidth]{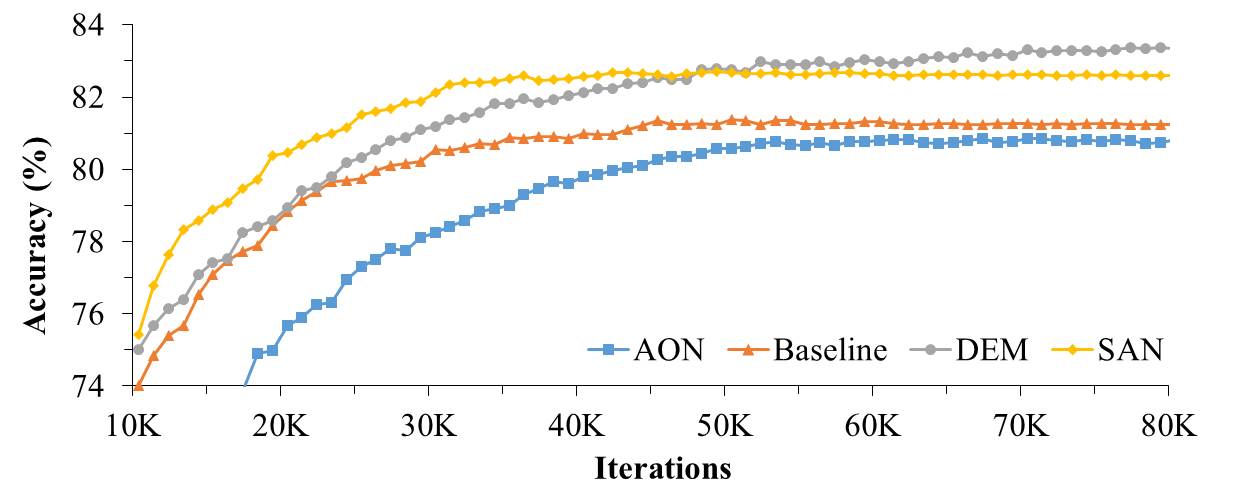}}
\caption{Comparison of the performance of the baseline and the proposed method. 
The x-axis represents the number of iterations and the y-axis represents the average accuracy of the test DB.
}
\label{various_models_results}
\end{figure}

\newcommand{\etal}{\textit{et al}.}
\begin{table*}[!h]
	\begin{center}
		\begin{tabular}{|l||c|c||c|c||c|c||c||c||c||c|}
			\hline
			\multirow{2}{*}{\textbf{Method}} &
			\multicolumn{2}{c|}{\textbf{IIIT5k}} & \multicolumn{2}{c|}{\textbf{SVT}} & \multicolumn{2}{c|}{\textbf{IC03}} & \textbf{IC13} & \textbf{IC15} & \textbf{SVTD} & \textbf{VTD}\cr\cline{2-11}
			& \textbf{50} & \textbf{None} & \textbf{50} & \textbf{None} & \textbf{50} & \textbf{None} & \textbf{None} & \textbf{None} & \textbf{None} & \textbf{None} \cr\hline
            ABBYY\cite{wang2011end} &24.3 & - & 35.0 & - & 56.0 & - & - & - & - & - \cr
            Wang \etal\cite{wang2011end} & - & - & 57.0 & - & 76.0 & - & - & - & - & - \cr
            Mishra \etal\cite{mishra2012scene} & 64.1 & - & 73.2 & - & 81.8 & - & - & - & - & - \cr
            Novikova \etal\cite{novikova2012large} & 64.1 & - & 72.9 & 82.8 & - & - & - & - & - & - \cr
            Wang \etal\cite{wang2012end} & - & - & 70.0 & - & 90.0 & - & - & - & - & - \cr
            Bissacco \etal\cite{bissacco2013photoocr} & - & - & 90.4 & - & - & - & 87.6 & - & - & - \cr
            Goel \etal\cite{goel2013whole} & - & - & 77.3 & - & 89.7 & - & - & - & - & - \cr
            Alsharif \etal\cite{alsharif2013end} & - & - & 74.3 & - & 93.1 & - & - & - & - & - \cr
            Almazan \etal\cite{almazan2014word} & 91.2 & - & 89.2 & - & - & - & - & - & - & - \cr
            Yao \etal\cite{yao2014strokelets} & 80.2 & - & 75.9 & - & 88.5 & - & - & - & - & - \cr
            Su and Lu  \etal\cite{su2014accurate} & - & - & 83.0 & - & 92.0 & - & - & - & - & - \cr
            Jaderberg \etal\cite{jaderberg2014deep} & 95.5 & - & 93.2 & 71.7 & 97.8 & 89.6 & 81.8 & - & - & - \cr
            Rodriguez \etal\cite{rodriguez2015label} & 76.1 & - & - & 70.0 & - & - & - & - & - & - \cr
            Gordo \etal\cite{gordo2015supervised} & 93.3 & - & 91.8 & - & - & - & - & - & - & - \cr
            Lee \etal\cite{lee2016recursive} & 96.8 & 78.4 & 96.3 & 80.7 & 97.9 & 88.7 & 90.0 & - & - & - \cr
			Shi \etal\cite{shi2017end} & 97.6 & 78.2 & 96.4 & 80.8 & 98.7 & 93.1 & 90.8 & - & - & - \cr
			Shi \etal\cite{Baoguang2016Robust} & 96.2 & 81.9 & 95.5 & 81.9 & 98.3 & 90.1 & 88.6 & - & - & - \cr
            Baoguang \etal\cite{Baoguang2016Robust} & 96.2 & 81.9 & 95.5 & 81.9 & 98.3 & 90.1 & 88.6 & - & - & - \cr
            Chen-Yu \etal\cite{lee2016recursive} & 96.8 & 78.4 & \textbf{96.3} & 80.7 & 97.9 & 88.7 & 90.0 & - & - & - \cr
			Cheng \etal\cite{Zhanzhan2017Arbitrarily} & \textbf{99.6} & \textbf{87.0} & 96.0 & 82.8 & 98.5 & 91.5 & 92.3 & 68.2 & - & - \cr\hline
            baseline & 96.5 & 82.3 & 93.3 & 83.4 & 96.5 & 93.1 & 91.5 & 67.9 & 87.0 & 62.0 \cr
            DEM & 97.5 & 83.3 & 93.8 & 83.6 & 98.1 & 93.5 & 92.7 & 69.7 & 87.9 & 65.2 \cr
            SAN & 97.1 & 83.0 & 93.7 & 85.5 & 98.5 & 93.3 & 92.5 & 68.0 & 87.6 & 64.8 \cr\hline
            ResNet+DEM & 98.6 & 85.8 & 94.1 & 86.7 & 98.3 & 94.0 & 93.7 & 73.1 & \textbf{90.5} & \textbf{70.9} \cr
			ResNet+SAN & 98.0 & 86.0 & 94.6 & \textbf{87.4} & \textbf{98.5} & \textbf{94.1} & \textbf{94.2} & \textbf{73.5} & 89.6 & 66.9 \cr\hline
		\end{tabular}
	\end{center}
	\caption{Results of recognition accuracy on general benchmarks. ``50'' are lexicon sizes, ``None'' means lexicon-free.}
	\label{tab:results}
\end{table*}

\subsection{Implementation Details}
The input images in our experiment are resized to $100\times32$ for VGG or $256\times32$ for ResNet\cite{Zhanzhan2017Focusing} and have one channel (gray-scale). The pixel values are normalized to the range (-0.5, 0.5) before they are provided as input to the network. All the LSTM cells used in our experiments have 512 memory units, and the dropouts are not used. The RMSProp method is used to conduct training with a batch size of 512. The initial learning rate is 0.001 and decreases by a factor of 0.9 as the validation errors stop decreasing for 10 iterations. Gradient clipping is applied at a magnitude of 5, and training is terminated after 100,000 iterations. To train our network, 8 million synthetic data released by\cite{JaderbergSVZ14} and 4 million synthetic instances (excluding images that contain non-alphanumeric characters) cropped from 90,000 images\cite{gupta2016synthetic} were used. In addition, we modified the open source by\cite{gupta2016synthetic} to generate 1 million synthetic vertical data for training and 3000 data for testing. We did not apply any data augmentation processes to the training dataset. Because we implemented our method using TensorFlow v1.6, CUDA9.0, and cuDNN v7.0, the proposed model can be GPU-accelerated. The experiments were performed on a workstation with one Intel Xeon(R) E5-2650 2.20GHz CPU and one NVIDIA Tesla M40 GPU.

\section{Results}
Fig.\ref{various_models_results} compares the performance of the baseline with that of the proposed method. As exhibited by the results, our baseline performed better than the AON-based method in terms of speed and accuracy. This is because an input image of 100 x 32 is effective while recognizing the horizontal and vertical characters that maintain an image input of 100 x 100. Additionally, it was better at recognizing characters while using the DEM and SAN methods of supervised learning than while using the baseline of unsupervised learning. In most of the benchmark datasets, DEM performed slightly better than SAN. 

\begin{figure}[!t]
\centering
\centerline{\includegraphics[width=1.0\linewidth]{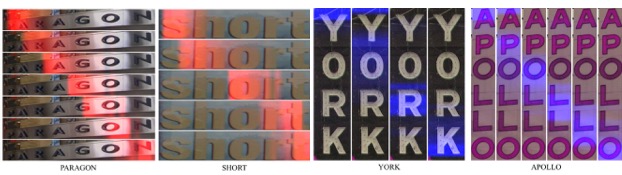}}
\caption{Results of attention and recognition for horizontal and vertical characters.
}
\label{result_attention}
\end{figure}

Table1 compares the performance of the state-of-the-art method and the proposed method. The combination of a ResNet-based CNN backbone exhibits the same effect as that exhibited by DEM and SAN. We achieved state-of-the-art performance for SVT, IC13, IC15, SVDT, and VDT142 based on a lexicon-free evaluation protocol. This exhibits that our algorithm works better in real world environments when there is no lexicon. Fig.\ref{result_attention} depicts the results of attention and recognition for horizontal and vertical characters, both of which confirm that the method works well. The main purpose of our study is to compare the performance between a supervised proposed method and an unsupervised baseline method for text direction. We would like to emphasize that our model not only recognizes vertical text but also recognizes horizontal text as compared with existing methods.

\section{Conclusions}
We presented a unified network to simultaneously deal with the horizontal and vertical texts. Our proposed DEM and SAN modules are supervised methods and exhibit better perceived performances than those exhibited by the previous methods. We achieved state-of-the-art results with respect to popular benchmark datasets and verified that our model works for a real vertical text dataset (VTD142).

\bibliographystyle{splncs}
\bibliography{main}

\end{document}